\documentclass[11pt,a4paper]{article}
\usepackage[T1]{fontenc}
\usepackage[hyperref]{acl2020}
\usepackage{times}
\usepackage{latexsym}

\usepackage{algorithm}
\usepackage{algpseudocode}
\usepackage{amsmath}
\usepackage{graphicx}

\usepackage{microtype}

\aclfinalcopy 

\title{TFW2V: An Enhanced Document Similarity Method for the Morphologically Rich Finnish Language}

\author{Quan Duong \\
  University of Helsinki / Finland \\
  \texttt{quan.duong@helsinki.fi} \\\And
  Mika Hämäläinen \\
  University of Helsinki / Finland \\
  \texttt{mika.hamalainen@helsinki.fi} \\\And
  Khalid Alnajjar  \\
  University of Helsinki / Finland \\
  \texttt{khalid.alnajjar@helsinki.fi} \\}

\author{Quan Duong,$^{1}$ Mika Hämäläinen,$^{1,2}$ Khalid Alnajjar$^{1,2}$\\
\tt{firstname.lastname@\{helsinki.fi\}}\\
$^{1}$University of Helsinki, $^{2}$Rootroo Ltd, Finland
}

\date{}

\begin{document}
\maketitle
\begin{abstract}
Measuring the semantic similarity of different texts has many important applications in Digital Humanities research such as information retrieval, document clustering and text summarization. The performance of different methods depends on the length of the text, the domain and the language. This study focuses on experimenting with some of the current approaches to Finnish, which is a morphologically rich language. At the same time, we propose a simple method, TFW2V, which shows high efficiency in handling both long text documents and limited amounts of data. Furthermore, we design an objective evaluation method which can be used as a framework for benchmarking text similarity approaches.
\end{abstract}

\section{Introduction}
\label{sec:intro}

Identifying documents that describe similar topics is a challenging yet important task. Detecting similar documents automatically has a wide range of digital humanities applications such as OCR post-correction \cite{dong-smith-2018-multi}, automatic clustering and linking of documents \cite{arnold-tilton-2018-cross,riedl-etal-2019-clustering} and clustering of semantic fields within a document \cite{35f79ff815cc4c3e8754a0a56121ab87}.

Assessing document similarity automatically becomes an important task especially due to the often unstructured nature of digital humanities research data (see \citealt{befd39df758e43fb87572aa4ace5037a}). This makes it possible to handle large text corpora in a more organized fashion by clustering similar texts together.

In this paper, we explore different approaches to textual similarity detection, namely TF-IDF, USE, Doc2Vec and our own proposed approach named TFW2V\footnote{Code available: https://github.com/ruathudo/tfw2v}. Our approach combines the traditional TF-IDF method with word embeddings to improve the overall performance of the text similarity method. Unlike the recent neural approaches such as BERT \cite{devlin2018bert}, RoBERTa \cite{liu2019roberta} or XLNet \cite{yang2019xlnet}, our method does not rely on a large external corpus, but can be fully trained on the texts the similarity of which one is to assess. This is useful since our model can then work on corpora that represent a different era than what modern NLP models are trained on, or even for languages that do not have massive text collections readily available or are spoken by communities that do not have access to the computational resources needed to train large neural language models.

\section{Related work}
\label{sec:related}
A survey conducted by \citet{Beel2016-11Resea-32348} showed that 83\% of text-based recommendation systems in digital libraries use TF–IDF. There is also a recent survey paper on the current state of Finnish NLP \cite{hamalainen2021current}. There is a number of papers studying automatic detection of genres \cite{dalan-sharoff-2016-genre,worsham-kalita-2018-genre,gianitsos-etal-2019-stylometric}, which, as a task, is not too far from ours. However, in this section, we focus mainly on approaches on document similarity.

\citet{KIM201915} have combined multiple document representation approaches, which are TF-IDF, LDA and Doc2Vec, to classify documents in a semi-supervised fashion. Their results indicate that combining the features of the aforementioned models enhanced the performance of the classification task. \citet{trucscua2019efficiency} has compared how different text representation models perform when training a Support Vector Machine (SVM) classifier. The results show that Doc2Vec was the superior model for the task addressed by the author, which is text categorization. \citet{duong2021benchmarks} also showed that clustering Finnish text is more effective by Doc2Vec compared to LDA.

A recent study by~\citet{marcinczuk-etal-2021-text} compared WordNet --a manually constructed network of concepts--, TF-IDF and word embeddings extracted from Doc2Vec and BERT for unsupervised classification of Polish text documents. Their study showed that manually constructed knowledge bases, i.e. WordNet in this case, is a valuable resource for the task. \citet{7774514} merged TF-IDF and Word Embeddings similarity scores to build the recommendation system for similar bug reports. 

\citet{li2019key} have used text representation models to extract keywords from short texts collected from social media by employing a TextRank~\cite{mihalcea2004textrank} algorithm which constructs a network and traverses it using random walk to discover the most important concepts. Text representation models have also been utilized with deep neural networks to classify text by~\citet{9425290}. 
TF-IDF and word embeddings have also been used to assess the similarity of entities \cite{hamalainen2021cute}. In particular, the authors used the aforementioned methods to extract and predict properties for Pokémon.

\section{Experiments}
\label{sec:experiments}
In this section, we apply four of the existing approaches to predict similarity of documents: Doc2Vec, Universal Sentence Encoder (USE), term frequency–inverse document frequency (TF-IDF) and average weighted word vector (AvgWV). Later on, we propose a new method to optimize TF-IDF by using a word embeddings model (TFW2V). All the experiments use the same datasets, the sampling process of which will be presented in the following section.

\subsection{Dataset}

We run the experiments based on the Yle News corpus. This corpus contains news articles published from 2011 to 2018 by the Finnish broadcasting company Yle (Yleisradio). The corpus is distributed through the Language Bank of Finland (Kielipankki)\footnote{\url{http://urn.fi/urn:nbn:fi:lb-2017070501}} and is freely available for research use\footnote{According to the license we cannot redistribute datasets derived from these data.}. There are more than 700,000 articles written in Finnish, each of which belongs to different categories with top-level categories such as Sport, Politics and Transportation. These categories have been defined by human authors and they have been coupled with keyword tags. For example, an article about a hockey match has the tags: \textit{urheilu} (sports), \textit{jääkiekon} (ice-hockey), \textit{miesten} (men's), \textit{sm-liiga} (The Finnish National League). The keyword tags illustrate well the contents of each article. 

Our study focuses on tackling the text similarity problem for documents as opposed to individual sentences or paragraphs. For this reason, we decided to filter the corpus to include only the articles that are between 200 and 600 words for the experiments. Next, we randomly sample 10 datasets from the filtered corpus so that each dataset contains 2000 unique articles. Thus, all datasets are independent from each other with no overlap. We only optimize the models for the first dataset as the training set. For testing, the models are applied to the rest of datasets with the extracted parameters without any modification.

\subsection{TF-IDF}
\label{sec:tfidf}
The first method we experiment with is TF-IDF. As stated in section \ref{sec:related}, this is a very simple method but it is very effective in many cases. The idea of this method can be expressed as follows: In a document, if a term (word) appears more frequently, it is given more weight, or a more important score. In contrast, if a term appears in many other documents in the corpus, it is regarded as less important or assumed to be a common word not descriptive enough for the document. The concurrence of these two metrics is combined in the equation below, to indicate the importance of a term in the text.

\begin{equation*}
\label{eq:tfidf}
W_{i,j} = TF_{i,j} \times \log(\frac{N}{DF_i}) 
\end{equation*}

In this equation, the $W_{i,j}$ is the weight of a term $i$ in document $j$, $TF_{i,j}$ is frequency of term $i$ in document $j$, $N$ is the number of documents in the corpus and $DF_i$ refers to the number of documents where the term {$i$} appears. The weights hence tend to filter out common terms and emphasize the important keywords of a given document. The value of TF-IDF weight is in range $[0, 1]$.

Before running the experiment, the text data is cleaned by removing punctuation and stopwords using NLTK \cite{bird2009natural}. For each sampled dataset, we calculate the TF-IDF weights for all documents. The pairs of terms and weights are feature vectors for each document. By using the cosine similarity function, we can measure the similarity between feature vectors. In order to not depend on magnitudes of vectors but their angles, this is a common metric to compute the semantic similarity for encoded text \citep{3320}. After having similarity scores calculated, they are saved for each pair of documents in dataset and sorted in descending order. From now, the top N similar documents can be queried from a given document.

\subsection{Average Weighted Word Vectors}
\label{sec:avg_wv}
Extending from the previous section \ref{sec:tfidf}, we introduce a combined method between TF-IDF and word embeddings algorithms called average weighted word vectors (AvgWV). This method was used in several previous researches to get a better representation for text document. \citet{RANI2021115867} used this method to get the representation of sentences in document to find the similarity between them. With the same approach, \citet{charbonnier-wartena-2018-using} applied to map the definition of an acronym with it context. 
The idea of this method is very easy to conduct. Both word embeddings and TF-IDF are trained for the given corpus. The representation of a document is the average of embedded vectors multiplied with the TF-IDF scores (weights) for all words in that document. By that, the TF-IDF scores punish the insignificant words and the influenced words have more impact on the averaged vector. The equation below is used to formulate the method.

\begin{equation*}
\label{eq:avg_wv}
\overrightarrow{D} = \frac{1}{N}\sum^{N}_{i} TF_i * \overrightarrow{WV}_i 
\end{equation*}

Where $\overrightarrow{D}$ is the vector representation of a document, $N$ is the number of word features. For each word $i$, we calculate the product of its TF-IDF score ($TF$) with its word vector ($\overrightarrow{WV}$) to get a new weighted vector. All weighted vectors corresponding to the word features are then averaged as the representation of document $D$.

The word embeddings model used in our experiment is based on Word2Vec from the work of \citet{mikolov2013efficient}. The model was trained in 20 epochs using the Gensim library with a vector size of 128, skip-gram method, negative windows of 5 for each sample dataset. Inherit from previous TF-IDF section \ref{sec:tfidf}, the averaged vectors are applied cosine distance to get the similarity score for documents. 

\subsection{Doc2Vec}
Documents originally stored in text format are convenient for humans to read, but they pose a challenge for computational tasks. Transforming from characters to a fixed length numeric representation is helpful for many purposes, for example: document retrieval, semantic comparison. One vector representation of text has been introduced by \citet{harris1954distributional} as a bag of words method. Even though this method is easy to compute and it shows the efficiency in many cases, there are still some weaknesses such as the lack of importance given to the word order and it suffers from data sparsity and high dimensionality. It is also missing the semantic meaning between the words, for example, the words ``dog'' and ``cat'' are more similar than ``dog'' and ``car'' but they are treated equally in the Bag of Words method.

Doc2Vec \citep{le2014distributed} is a document embeddings algorithm that comes to solve the issue from Bag of Words. The advantage of Doc2Vec is to vectorize a whole text document regardless of its length and to provide the semantic relationship of words. 

We use the existing implementation of Doc2Vec in Gensim library \cite{rehurek_lrec}. Before training, the text is tokenized and all stopwords are removed using NLTK. The setup on Doc2Vec model is kept in default with a dimensionality to 100 for vector size, negative sampling of 5 words and train for 30 epochs. We train model for each dataset separately. Thus there are 10 different Doc2Vec models corresponding to the datasets. Cosine similarity is again applied to these dense document vectors from the Doc2Vec model to get the similarity scores between documents.

\subsection{Universal Sentence Encoder}
The next approach we experienced is using the Universal Sentence Encoder (USE) \citep{yang2019multilingual} model for multi-languages. The USE model was trained based on the Transformer architecture \cite{vaswani2017attention} for over 16 languages which shows a very good performance for various semantic textual similarity tasks. However, this model does not support Finnish. 

\citet{reimers-2020-multilingual-sentence-bert} introduced a novel way to transfer knowledge of a sentence encoder model from one language to another. 
On that paper, DistilmBERT \cite{Sanh2019DistilBERTAD} model, a distilled version of BERT \cite{devlin2018bert} trained on 104 different languages \footnote{Provided through the Transformers Python library \cite{wolf2019huggingface} \url{https://huggingface.co/distilbert-base-multilingual-cased}}, was selected as student model. It is then adapted to USE model \citep{yang2019multilingual} (as a teacher model) to support 50+ languages including Finnish. The pre-trained model was published with the name ``distiluse-base-multilingual-cased-v2'' in the Sentence-Transformers library \citep{reimers-2019-sentence-bert}. 

We applied the pre-trained model without any modifications. The maximum length support for the text is 512 tokens. The whole document is encoded automatically by the model and output as a dense vector. With the collected vectors we are able to compare the similarity between documents using cosine similarity.


\subsection{Enrich TF-IDF by Word Embeddings (TFW2V)}

\begin{algorithm*}[ht]
\caption{Enrich TF-IDF}
\begin{algorithmic}
\Procedure{EnrichTFIDF}{$Features, SimDocs, W2V, MinWeight, MaxTerm, Alpha$}

    \State \texttt{\textbf{sort} $Features$(TermID, Weight) in DESC order of Weight}
    \State \texttt{\textbf{filter} $Features$ have $Weight$ < $MinWeight$}
    \State \texttt{\textbf{trim} $Features$ to $MaxTerm$ \textbf{if} length $Features$ > $MaxTerm$}
    
    \For{\texttt{$SimFeatures, SimScore$ in $SimDocs$}}

        \State \texttt{\textbf{sort} $SimFeatures$(TermID, Weight) in DESC order of Weight}
        \State \texttt{\textbf{filter} $SimFeatures$ have $Weight$ < $MinWeight$}
        \State \texttt{\textbf{trim} $SimFeatures$ to $MaxTerm$ \textbf{if} length $SimFeatures$ > $MaxTerm$}
        
        \State $WVScore$ = $W2V.calculate\_similarity(Features, SimFeatures)$
        \State $NewScore$ = $(WVScore \times Alpha + SimScore) / (1 + Alpha)$
        \State \texttt{\textbf{save} $NewScore$}
    \EndFor
\EndProcedure

\end{algorithmic}
\label{algo:tfw2v}
\end{algorithm*}

The following part of this paper moves on to describe our modified version of TF-IDF algorithm. As introduced in section TF-IDF \ref{sec:tfidf}, this algorithm is very simple to compare similarity of documents. However, it also has many drawbacks. Firstly, the position of words in text is completely ignored. Secondly, because of relying on the lexical features, it skips semantic relationship of words. For example, with the synonyms or plural form of words, TF-IDF treats them as separated features without any linking. This will have a huge impact on morphologically rich languages such as Finnish, which contains many inflectional forms for all words and their compounds (see \citealt{duong-etal-2021-unsupervised}) even when lematization is applied. To overcome the issues, we propose a new algorithm that uses a word embeddings model to enrich the TF-IDF result.
The details of the algorithm are presented in pseudo code \ref{algo:tfw2v}.

\begin{figure}[h]
\centering
\includegraphics[width=0.45\textwidth]{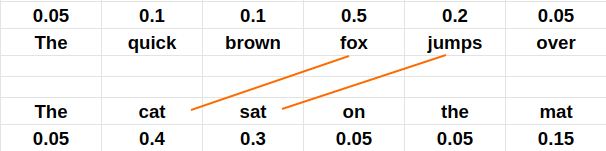}
\caption{The two texts have the words \textbf{Fox} and \textbf{Cat} with high TF-IDF weights. At the same time, they have semantic similarity in the Word2Vec model, so that the documents can be linked. Same is applied to the words \textbf{Jumps} and \textbf{Sat}}
\label{fig:fox_cat}
\end{figure}

The general idea of the algorithm can be explained as follows. We train a word embeddings model from the same corpus, so the words or terms of documents have semantic relationships. The word embeddings model can be used to measure the similarity of two terms. Turning now to TF-IDF output, the terms or features of a document contain the important information with higher weights. These important terms of two documents can be semantically linked by using a trained word embeddings model. An example is shown in the figure \ref{fig:fox_cat} to better explain.

The level of similarity between two group features is used to give additional reward on the final similarity score between a pair of documents. For example, if document A has important features (term1, term3, term8) and document B has important features (term2, term5, term9), the similarity score between document A and B can be added a small portion from the semantic similarity score between two features group. Similar to AvgWV section \ref{sec:avg_wv}, the Word2Vec (W2V) model was used for word embeddings. The model was trained in 20 epochs using the Gensim library with a vector size of 128, skip-gram method, negative windows of 5 for each sample dataset.

To determine how much reward should be added to the TF-IDF similarity score, we design three parameters: \textbf{MinWeight}, \textbf{MaxTerm} and \textbf{Alpha}. Let take a look at the algorithm \ref{algo:tfw2v}. Given a list of features (terms with weights) from a document and a list of similar documents as the result from TF-IDF, we want to change the result or re-rank it. Firstly, the features are sorted in the descending order of weight. The \textbf{MinWeight} parameter is used to filter important features, higher it is, less features are kept for comparison (lower bound). In some cases, the number of features considered as essential is too high, and we want to trim them to a certain number by the \textbf{MaxTerm} number (upper bound). For the list of similar documents, we apply the same process. After that, we get the similarity score between given features and compared features by W2V model. Note that, the W2V model generated by Gensim provides a method to compute similarity score of two set of words by averaging vectors for each set \footnote{\url{https://radimrehurek.com/gensim/models/keyedvectors.html\#gensim.models.keyedvectors.KeyedVectors.n_similarity}}. Next, the new similarity score is calculated by the following formula:  
\begin{equation*}
    NewScore = \frac{WVScore \times Alpha + SimScore}{1 + Alpha}
\end{equation*}

Where the \textbf{SimScore} is the similarity score from TF-IDF, \textbf{WVScore} is the similarity score from W2V model for the important features, and \textbf{Alpha} is the parameter to decide how W2V similarity influences the current score. When \textbf{Alpha} is equal to 0, it has no effect, and when it is set to 1, the new score is the average of the two scores. Larger \textbf{Alpha} will have a higher recall which is bound to link an increasing number of unexpected documents together, while a smaller number yields a more conservative end result. In our experiments, we empirically set $MinWeight = 0.08$, $MaxLength = 20$ and $Alpha = 0.1$. These parameters are consistent for all datasets. Finally, the results are re-ranked for the new similarity scores.

\begin{figure*}[ht]
\centering
\includegraphics[width=\textwidth]{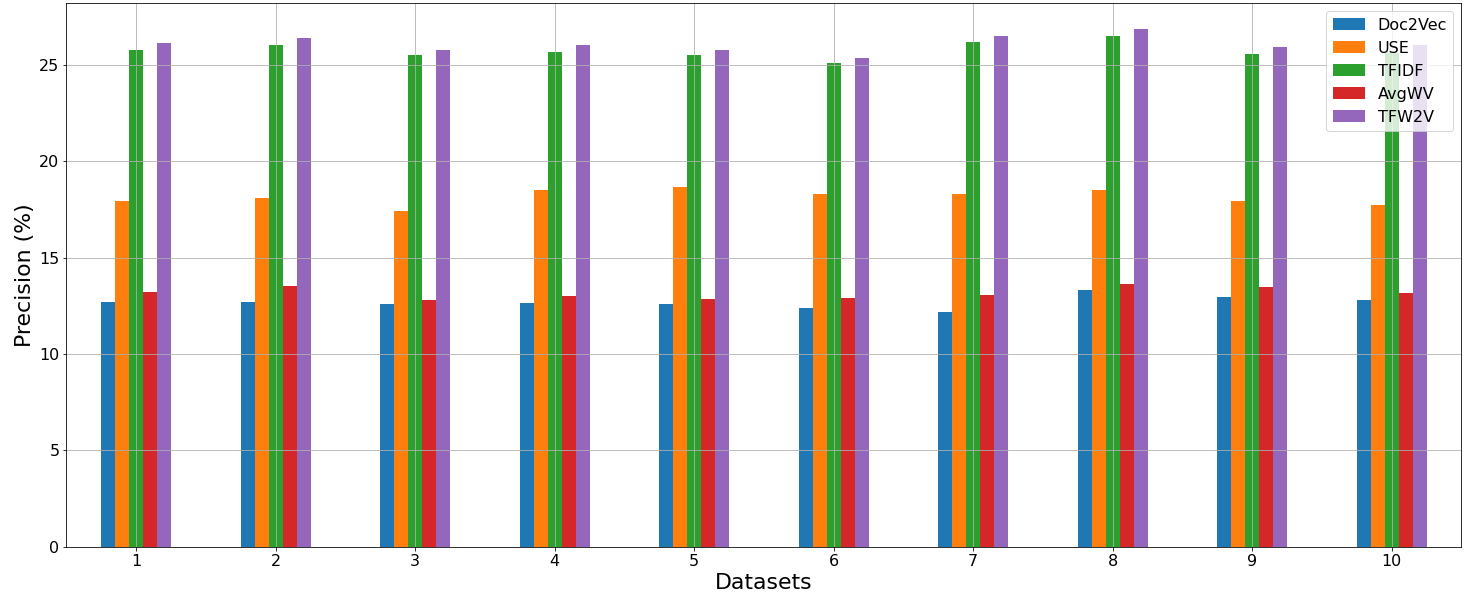}
\caption{Precision accuracy for all datasets on Top-30 similarity. Higher is better.}
\label{fig:barplot}
\end{figure*}

\begin{figure*}[ht]
\centering
\includegraphics[width=\textwidth]{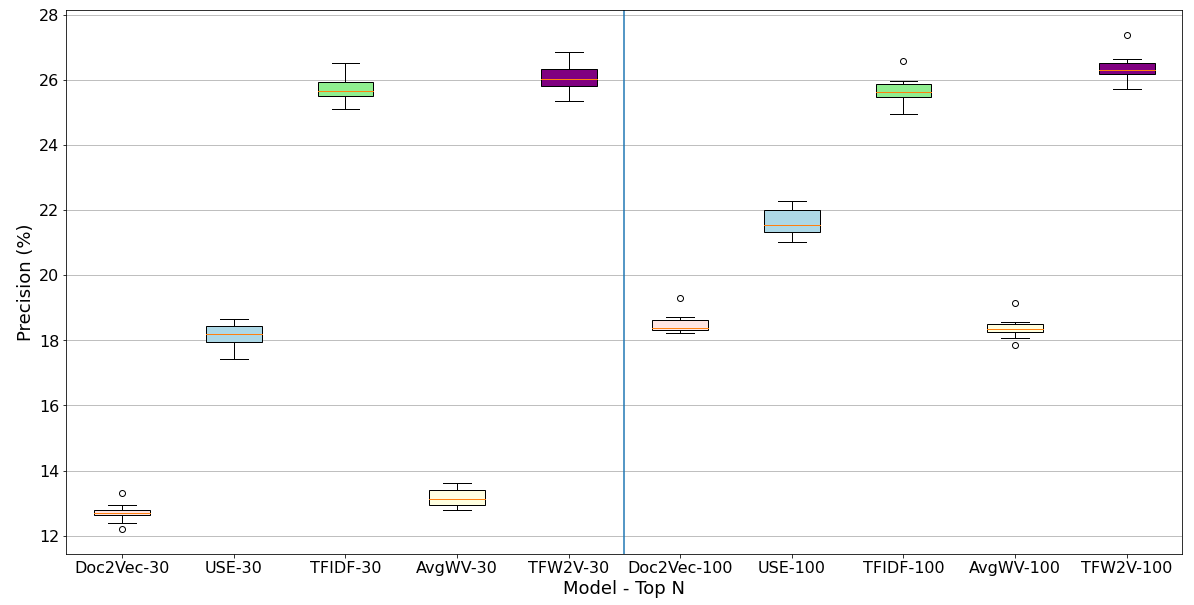}
\caption{Precision accuracy for all datasets on Top-30 and Top-100 similarity. Higher is better.}
\label{fig:boxplot_precision}
\end{figure*}

\section{Evaluation}
\label{sec:evaluation}
Turning now to the evaluation, as previously stated, we have 10 independent datasets for benchmarking. All the experimented parameters from the models are applied consistently for those datasets. We assess the performance of the 5 methods TF-IDF, AvgWV, Doc2Vec, USE and TFW2V by three criteria: Top-N Precision, Top-N BLEU score and Top-N ranking loss. We will explain those metrics in the following sections of the paper.

\begin{figure*}[ht]
\centering
\includegraphics[width=\textwidth]{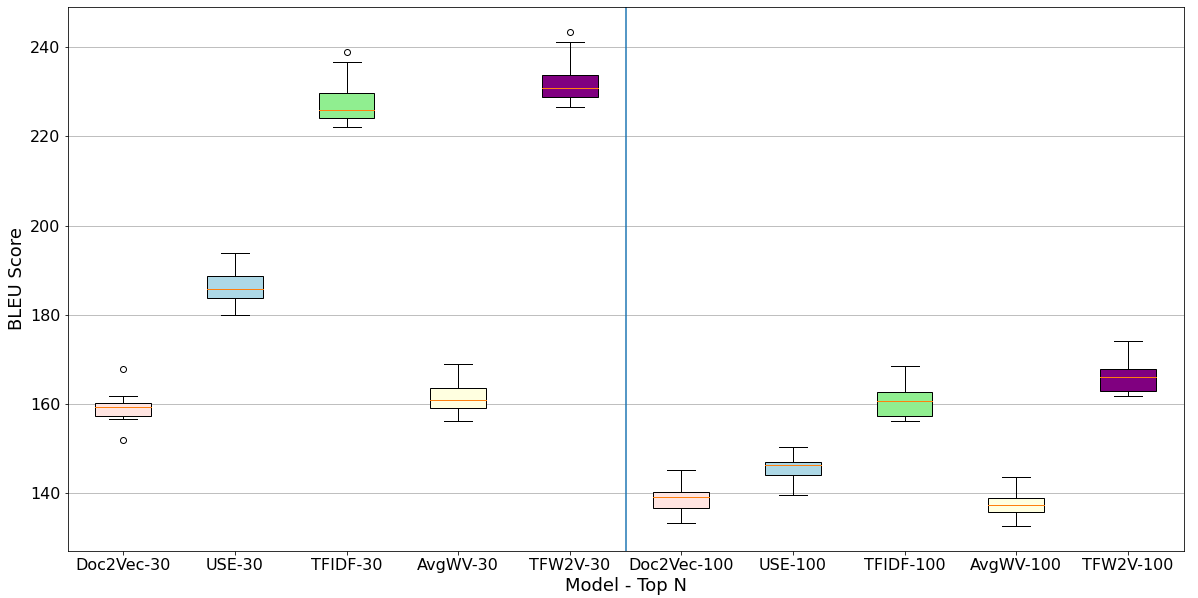}
\caption{Sum of BLEU scores for 2k documents in each dataset. All datasets are evaluated for Top-30 and Top-100 similarity. Higher is better.}
\label{fig:boxplot_bleu}
\end{figure*}

\subsection{Ground Truth}
\label{sec:groundtruth}
The ground truth for evaluation is created by the tags attached to articles. Because the tags are manually labeled by human authors to illustrate the content of articles, comparing the similarity between sets of tags can reflect the similarity of articles. There are many ways to measure similarity of two sets, such as counting overlapping tags. In machine translation, BLEU score \citep{papineni-etal-2002-bleu} is a popular method to evaluate the translated sentence quality. BLEU method calculates the similarity between two sets of words, very close to our case. The difference is only that two sentences in machine translation have N-grams dependence while similarity of two sets of tags are not relied on the position of tags. We use a simplified version of BLEU score, which is calculating score for unigrams without considering other higher order N-grams. After having BLEU scores for all document pairs, we sort them in descending order for evaluation.

\begin{figure*}[ht]
\centering
\includegraphics[width=\textwidth]{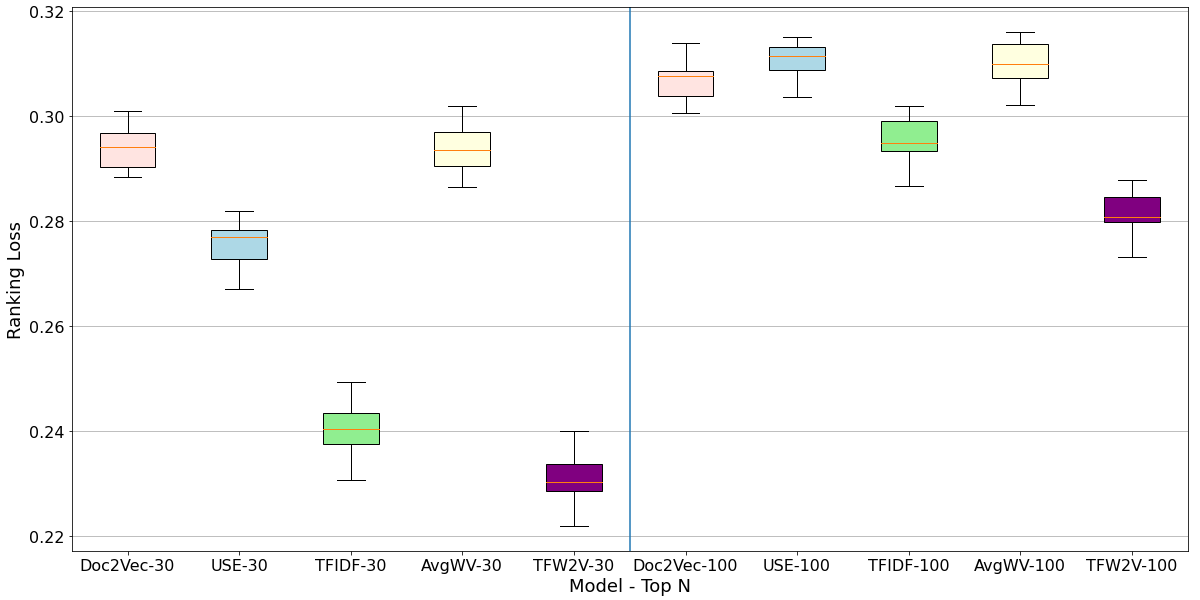}
\caption{MAE Ranking loss for all datasets on Top-30 and Top-100 similarity. Lower is better.}
\label{fig:boxplot_mae}
\end{figure*}

\subsection{Top-N Precision}

The first metric is Top-N Precision. The metric can be presented as in the Top-N documents predicted as the most similar to a given one, i.e. how many documents are correctly ranked. For instance, given a document with top 100 similar documents, there are 40 documents that are ranked correctly in the top 100, the precision for it is 40\%. The precision is calculated for all 2000 documents and averaged for each dataset. The formula below is to calculate the precision, where the $D_{pred}$ is a set of predicted documents, $D_{real}$ is a set of ground truth, and N is the number of documents for Top-N:

\begin{equation*}
    Precision = \frac{\sum {D_{pred} \cap D_{real}}}{N}
\end{equation*}

In the figure \ref{fig:barplot}, the precision for Top-30 is presented. For all the 10 datasets in figure \ref{fig:barplot}, the TFW2V model outperforms the other models clearly with an \textbf{26.09\%} average (avg) accuracy. The Doc2Vec model has the lowest accuracy (\textbf{12.70\%} avg). While the USE model shows much better results compared to Doc2Vec (\textbf{18.15\%} avg), it is still inferior to TF-IDF (\textbf{25.76\%} avg). The AvgWV is approximately comparable with Doc2Vec with slightly better numbers (\textbf{13.17\%} avg). Similar results also take place for the Top-100, where the TFW2V surpassed the other models in every dataset. Therefore, to have a better visualization, we use boxplot to illustrate not only the difference between models, but also the Top-N variants.

The figure \ref{fig:boxplot_precision} demonstrates the precision for both Top-30 and Top-100 results. It is interesting that the results from Doc2Vec (\textbf{18.51\%} avg), AvgWV (\textbf{18.38\%} avg) and USE (\textbf{21.64\%} avg) are significantly improved for the Top-100. This can be explained as the more relaxing the boundary, the higher the chance a document is predicted correctly in the Top-N list. However, the TF-IDF result has not improved. Despite depending on the TF-IDF results, TFW2V-100 still shows a slight increase (\textbf{26.41\%} avg) compared to Top-30 results.

\subsection{Top-N BLEU score}
The next metric is Top-N BLEU score to measure how relevant a group of similar documents is to a given document. The way to conduct this metric is very similar to calculating the BLEU score for ranking in the Ground Truth section \ref{sec:groundtruth}. We calculate the BLEU score on a unigram level for the tags of a given document against the Top-N similar documents tags. These scores are then averaged for N similar documents. From there, we sum all the averaged BLEU scores for 2000 documents in a dataset. The averaged BLEU score is calculated as follow:

\begin{equation*}
    Score = \frac{\sum^{N}_{i=1} BLEU(T, T_i)}{N}
\end{equation*}

where $T$ is the tags from a given document and $T_i$ is the tags from similar documents and N is the number of Top-N. 
From the results we observed, the TFW2V model again outperforms the other models in all datasets. The boxplot in figure \ref{fig:boxplot_bleu} shows the precise expression of performances for all models. We can see, Doc2Vec still remains less effective, around \textbf{160} for Top-30 and \textbf{139} for Top-100 on average. Similar numbers to Doc2Vec come from AvgWV method. The USE model results (\textbf{186} and \textbf{145}) are still lower compared to TF-IDF (\textbf{228} and \textbf{160}). Our proposed model TFW2V showed the improvement to TF-IDF with \textbf{233} and \textbf{166} scores on both Top-30 and Top-100 respectively.

The overall result for Top-30 is higher than Top-100. This is understandable as the more documents in Top-N there are, the more irrelevant ones making to the list will make the average scores decrease.

\subsection{Top-N Ranking loss}

The final metric we want to introduce is Top-N Ranking loss. This metric reflects how far a predicted position of similar documents is to the real order in ground truth for the Top-N. For example, in Top-30, we compare 30 predicted document orders to their real orders. If a predicted document has the position 5 and its real position is 45, the loss between the two orders is 40. Thus, the average loss for all documents is calculated using the Mean Absolute Error (MAE) function. The MAE loss is then divided for the length of the dataset (length of max rank) for normalization. The formula below is for calculating MAE loss between two positions $P$ (ground truth) and $\hat{P}$ (prediction) for each document in Top-N with $S$ is the length of the dataset.

\begin{equation*}
    Loss = \frac{\sum^{N}_{i=1} |P_i - \hat{P_i}|}{N \times S}
\end{equation*}

We got the result for this metric illustrated in figure \ref{fig:boxplot_mae}. This time, both Doc2Vec and AvgWV models show the highest loss at Top-30 with loss around \textbf{0.29}. Interestingly, they are a bit better than the USE model at Top-100 (\textbf{0.30} vs \textbf{0.31}). TF-IDF is still obviously impressive compared to the previous ones with \textbf{0.24} for Top-30 and \textbf{0.29} for Top-100. Though, TFW2V is continuing to achieve the best result with the lowest losses of \textbf{0.23} and \textbf{0.28} for Top-30 and Top-100 respectively. This also indicates that the TFW2V model gives less irrelevant documents in Top-N than the other models.

\section{Conclusion}
\label{sec:conclusion}

In summary, we have presented a simple method to improve the TF-IDF algorithm by using a word embeddings model. The proposed method outperforms the more complex models like Doc2Vec and USE. We also compared it to a popular method AvgWV which use the same combination of TF-IDF and Word2Vec but in different way. It is very obvious that our proposed approach is surpassing the AvgWV model. The weakness of AvgWV is that it's hard to control the averaged vector representation of a document when all words and their TF-IDF weights are taken into account. Additionally, the impact from word vectors could come to too strong in some cases, which create the bias in the final decision. 

In our method TFW2V, we can control the effect of word embedding model on the similarity score. At the same time, not all the words are considered into the enhancing process but the important ones. Thus, it is more stable, flexible and controllable to apply in various purposes. For example, in the document retriever system, the parameters can be set to get the relevant result as priority. On the other hand, in a recommender system, the parameters can be adjusted to get more creative result, thus it can look up for the under-discovered articles. 

It is clearly observable that with a morphologically rich language like Finnish, TF-IDF still works very effectively. However, by combining it with a Word2Vec model and our algorithm \ref{algo:tfw2v}, the result is significantly enhanced. The method is entirely unsupervised and works well with a small dataset like in our experiment with only 2000 samples. 

In the future work, we will experiment this method for more languages and different lengths of document. The source code of this project will be provided as a Python library\footnote{\url{https://github.com/ruathudo/tfw2v}} which is easy to install and apply for any DH research. The lack of dependency on neural language models trained on massive amounts of data makes our approach applicable in scenarios where such amounts of text are unfeasible to obtain.



\bibliography{anthology,acl2020}

\begin{thebibliography}{38}
\expandafter\ifx\csname natexlab\endcsname\relax\def\natexlab#1{#1}\fi

\bibitem[{Arnold and Tilton(2018)}]{arnold-tilton-2018-cross}
Taylor Arnold and Lauren Tilton. 2018.
\newblock \href {https://www.aclweb.org/anthology/W18-4506} {Cross-discourse
  and multilingual exploration of textual corpora with the {D}ual{N}eighbors
  algorithm}.
\newblock In \emph{Proceedings of the Second Joint {SIGHUM} Workshop on
  Computational Linguistics for Cultural Heritage, Social Sciences, Humanities
  and Literature}, pages 50--59, Santa Fe, New Mexico. Association for
  Computational Linguistics.

\bibitem[{Beel et~al.(2016)Beel, Gipp, Langer, and
  Breitinger}]{Beel2016-11Resea-32348}
Joeran Beel, Bela Gipp, Stefan Langer, and Corinna Breitinger. 2016.
\newblock \href {https://doi.org/10.1007/s00799-015-0156-0} {Research-paper
  recommender systems : a literature survey}.
\newblock \emph{International Journal on Digital Libraries}, 17(4):305--338.

\bibitem[{Bird et~al.(2009)Bird, Klein, and Loper}]{bird2009natural}
Steven Bird, Ewan Klein, and Edward Loper. 2009.
\newblock \emph{Natural language processing with Python: analyzing text with
  the natural language toolkit}.
\newblock " O'Reilly Media, Inc.".

\bibitem[{Charbonnier and Wartena(2018)}]{charbonnier-wartena-2018-using}
Jean Charbonnier and Christian Wartena. 2018.
\newblock \href {https://www.aclweb.org/anthology/C18-1221} {Using word
  embeddings for unsupervised acronym disambiguation}.
\newblock In \emph{Proceedings of the 27th International Conference on
  Computational Linguistics}, pages 2610--2619, Santa Fe, New Mexico, USA.
  Association for Computational Linguistics.

\bibitem[{Dalan and Sharoff(2016)}]{dalan-sharoff-2016-genre}
Erika Dalan and Serge Sharoff. 2016.
\newblock \href {https://doi.org/10.18653/v1/W16-2611} {Genre classification
  for a corpus of academic webpages}.
\newblock In \emph{Proceedings of the 10th Web as Corpus Workshop}, pages
  90--98, Berlin. Association for Computational Linguistics.

\bibitem[{Devlin et~al.(2018)Devlin, Chang, Lee, and
  Toutanova}]{devlin2018bert}
Jacob Devlin, Ming-Wei Chang, Kenton Lee, and Kristina Toutanova. 2018.
\newblock \href {http://arxiv.org/abs/1810.04805} {{BERT}: Pre-training of deep
  bidirectional transformers for language understanding}.

\bibitem[{Dogru et~al.(2021)Dogru, Tilki, Jamil, and Ali~Hameed}]{9425290}
Hasibe~Busra Dogru, Sahra Tilki, Akhtar Jamil, and Alaa Ali~Hameed. 2021.
\newblock \href {https://doi.org/10.1109/CAIDA51941.2021.9425290} {Deep
  learning-based classification of news texts using doc2vec model}.
\newblock In \emph{2021 1st International Conference on Artificial Intelligence
  and Data Analytics (CAIDA)}, pages 91--96.

\bibitem[{Dong and Smith(2018)}]{dong-smith-2018-multi}
Rui Dong and David Smith. 2018.
\newblock \href {https://doi.org/10.18653/v1/P18-1220} {Multi-input attention
  for unsupervised {OCR} correction}.
\newblock In \emph{Proceedings of the 56th Annual Meeting of the Association
  for Computational Linguistics (Volume 1: Long Papers)}, pages 2363--2372,
  Melbourne, Australia. Association for Computational Linguistics.

\bibitem[{Duong et~al.(2021{\natexlab{a}})Duong, H{\"a}m{\"a}l{\"a}inen, and
  Hengchen}]{duong-etal-2021-unsupervised}
Quan Duong, Mika H{\"a}m{\"a}l{\"a}inen, and Simon Hengchen.
  2021{\natexlab{a}}.
\newblock \href {https://aclanthology.org/2021.nodalida-main.24} {An
  unsupervised method for {OCR} post-correction and spelling normalisation for
  {F}innish}.
\newblock In \emph{Proceedings of the 23rd Nordic Conference on Computational
  Linguistics (NoDaLiDa)}, pages 240--248, Reykjavik, Iceland (Online).
  Link{\"o}ping University Electronic Press, Sweden.

\bibitem[{Duong et~al.(2021{\natexlab{b}})Duong, Pivovarova, and
  Zosa}]{duong2021benchmarks}
Quan Duong, Lidia Pivovarova, and Elaine Zosa. 2021{\natexlab{b}}.
\newblock \href {http://ceur-ws.org/Vol-2981/paper5.pdf} {Benchmarks for
  unsupervised discourse change detection}.
\newblock In \emph{Proceedings of the 6th International Workshop on
  Computational History}.

\bibitem[{Gianitsos et~al.(2019)Gianitsos, Bolt, Chaudhuri, and
  Dexter}]{gianitsos-etal-2019-stylometric}
Efthimios Gianitsos, Thomas Bolt, Pramit Chaudhuri, and Joseph Dexter. 2019.
\newblock \href {https://doi.org/10.18653/v1/W19-2507} {Stylometric
  classification of ancient {G}reek literary texts by genre}.
\newblock In \emph{Proceedings of the 3rd Joint {SIGHUM} Workshop on
  Computational Linguistics for Cultural Heritage, Social Sciences, Humanities
  and Literature}, pages 52--60, Minneapolis, USA. Association for
  Computational Linguistics.

\bibitem[{H{\"a}m{\"a}l{\"a}inen and
  Alnajjar(2019)}]{35f79ff815cc4c3e8754a0a56121ab87}
Mika H{\"a}m{\"a}l{\"a}inen and Khalid Alnajjar. 2019.
\newblock \href {https://doi.org/10.18653/v1/w19-8637} {Let{\textquoteright}s
  face it: Finnish poetry generation with aesthetics and framing}.
\newblock In \emph{12th International Conference on Natural Language
  Generation}, pages 290--300, United States. The Association for Computational
  Linguistics.

\bibitem[{H{\"a}m{\"a}l{\"a}inen and Alnajjar(2021)}]{hamalainen2021current}
Mika H{\"a}m{\"a}l{\"a}inen and Khalid Alnajjar. 2021.
\newblock The current state of {F}innish {NLP}.
\newblock In \emph{Proceedings of the Seventh International Workshop on
  Computational Linguistics of Uralic Languages}, pages 65--72.

\bibitem[{H{\"a}m{\"a}l{\"a}inen et~al.(2021)H{\"a}m{\"a}l{\"a}inen, Alnajjar,
  and Partanen}]{hamalainen2021cute}
Mika H{\"a}m{\"a}l{\"a}inen, Khalid Alnajjar, and Niko Partanen. 2021.
\newblock How cute is {P}ikachu? gathering and ranking {P}okémon properties
  from data with {P}okémon word embeddings.
\newblock \emph{arXiv preprint arXiv:2108.09546}.

\bibitem[{Harris(1954)}]{harris1954distributional}
Zellig~S Harris. 1954.
\newblock Distributional structure.
\newblock \emph{Word}, 10(2-3):146--162.

\bibitem[{Kim et~al.(2019)Kim, Seo, Cho, and Kang}]{KIM201915}
Donghwa Kim, Deokseong Seo, Suhyoun Cho, and Pilsung Kang. 2019.
\newblock \href {https://doi.org/https://doi.org/10.1016/j.ins.2018.10.006}
  {Multi-co-training for document classification using various document
  representations: Tf–idf, lda, and doc2vec}.
\newblock \emph{Information Sciences}, 477:15--29.

\bibitem[{Le and Mikolov(2014)}]{le2014distributed}
Quoc Le and Tomas Mikolov. 2014.
\newblock Distributed representations of sentences and documents.
\newblock In \emph{International conference on machine learning}, pages
  1188--1196.

\bibitem[{Li et~al.(2019)Li, Huang, Fan, Sun, and Zhu}]{li2019key}
Jun Li, Guimin Huang, Chunli Fan, Zhenglin Sun, and Hongtao Zhu. 2019.
\newblock Key word extraction for short text via word2vec, doc2vec, and
  textrank.
\newblock \emph{Turkish Journal of Electrical Engineering \& Computer
  Sciences}, 27(3):1794--1805.

\bibitem[{Liu et~al.(2019)Liu, Ott, Goyal, Du, Joshi, Chen, Levy, Lewis,
  Zettlemoyer, and Stoyanov}]{liu2019roberta}
Yinhan Liu, Myle Ott, Naman Goyal, Jingfei Du, Mandar Joshi, Danqi Chen, Omer
  Levy, Mike Lewis, Luke Zettlemoyer, and Veselin Stoyanov. 2019.
\newblock Roberta: A robustly optimized bert pretraining approach.
\newblock \emph{arXiv preprint arXiv:1907.11692}.

\bibitem[{M{\"a}kel{\"a} et~al.(2020)M{\"a}kel{\"a}, Lagus, Lahti, S{\"a}ily,
  Tolonen, H{\"a}m{\"a}l{\"a}inen, Kaislaniemi, and
  Nevalainen}]{befd39df758e43fb87572aa4ace5037a}
Eetu M{\"a}kel{\"a}, Krista Lagus, Leo Lahti, Tanja S{\"a}ily, Mikko Tolonen,
  Mika H{\"a}m{\"a}l{\"a}inen, Samuli Kaislaniemi, and Terttu Nevalainen. 2020.
\newblock Wrangling with non-standard data.
\newblock In \emph{Proceedings of the Digital Humanities in the Nordic
  Countries 5th Conference}, number 2612 in CEUR Workshop Proceedings, pages
  81--96, Germany. CEUR-WS.org.

\bibitem[{Marci{\'n}czuk et~al.(2021)Marci{\'n}czuk, Gniewkowski, Walkowiak,
  and B{\k{e}}dkowski}]{marcinczuk-etal-2021-text}
Micha{\l} Marci{\'n}czuk, Mateusz Gniewkowski, Tomasz Walkowiak, and Marcin
  B{\k{e}}dkowski. 2021.
\newblock \href {https://aclanthology.org/2021.gwc-1.24} {Text document
  clustering: {W}ordnet vs. {TF}-{IDF} vs. word embeddings}.
\newblock In \emph{Proceedings of the 11th Global Wordnet Conference}, pages
  207--214, University of South Africa (UNISA). Global Wordnet Association.

\bibitem[{Mihalcea and Tarau(2004)}]{mihalcea2004textrank}
Rada Mihalcea and Paul Tarau. 2004.
\newblock Textrank: Bringing order into text.
\newblock In \emph{Proceedings of the 2004 conference on empirical methods in
  natural language processing}, pages 404--411.

\bibitem[{Mikolov et~al.(2013)Mikolov, Chen, Corrado, and
  Dean}]{mikolov2013efficient}
Tomas Mikolov, Kai Chen, Greg Corrado, and Jeffrey Dean. 2013.
\newblock Efficient estimation of word representations in vector space.
\newblock \emph{arXiv preprint arXiv:1301.3781}.

\bibitem[{Papineni et~al.(2002)Papineni, Roukos, Ward, and
  Zhu}]{papineni-etal-2002-bleu}
Kishore Papineni, Salim Roukos, Todd Ward, and Wei-Jing Zhu. 2002.
\newblock \href {https://doi.org/10.3115/1073083.1073135} {{B}leu: a method for
  automatic evaluation of machine translation}.
\newblock In \emph{Proceedings of the 40th Annual Meeting of the Association
  for Computational Linguistics}, pages 311--318, Philadelphia, Pennsylvania,
  USA. Association for Computational Linguistics.

\bibitem[{Rani and Lobiyal(2021)}]{RANI2021115867}
Ruby Rani and Daya~K. Lobiyal. 2021.
\newblock \href {https://doi.org/https://doi.org/10.1016/j.eswa.2021.115867} {A
  weighted word embedding based approach for extractive text summarization}.
\newblock \emph{Expert Systems with Applications}, 186:115867.

\bibitem[{{\v R}eh{\r u}{\v r}ek and Sojka(2010)}]{rehurek_lrec}
Radim {\v R}eh{\r u}{\v r}ek and Petr Sojka. 2010.
\newblock {Software Framework for Topic Modelling with Large Corpora}.
\newblock In \emph{{Proceedings of the LREC 2010 Workshop on New Challenges for
  NLP Frameworks}}, pages 45--50, Valletta, Malta. ELRA.
\newblock \url{http://is.muni.cz/publication/884893/en}.

\bibitem[{Reimers and Gurevych(2019)}]{reimers-2019-sentence-bert}
Nils Reimers and Iryna Gurevych. 2019.
\newblock \href {http://arxiv.org/abs/1908.10084} {Sentence-bert: Sentence
  embeddings using siamese bert-networks}.
\newblock In \emph{Proceedings of the 2019 Conference on Empirical Methods in
  Natural Language Processing}. Association for Computational Linguistics.

\bibitem[{Reimers and Gurevych(2020)}]{reimers-2020-multilingual-sentence-bert}
Nils Reimers and Iryna Gurevych. 2020.
\newblock \href {http://arxiv.org/abs/2004.09813} {Making monolingual sentence
  embeddings multilingual using knowledge distillation}.
\newblock \emph{arXiv preprint arXiv:2004.09813}.

\bibitem[{Riedl et~al.(2019)Riedl, Betz, and
  Pad{\'o}}]{riedl-etal-2019-clustering}
Martin Riedl, Daniela Betz, and Sebastian Pad{\'o}. 2019.
\newblock \href {https://doi.org/10.18653/v1/W19-2502} {Clustering-based
  article identification in historical newspapers}.
\newblock In \emph{Proceedings of the 3rd Joint {SIGHUM} Workshop on
  Computational Linguistics for Cultural Heritage, Social Sciences, Humanities
  and Literature}, pages 12--17, Minneapolis, USA. Association for
  Computational Linguistics.

\bibitem[{Sanh et~al.(2019)Sanh, Debut, Chaumond, and
  Wolf}]{Sanh2019DistilBERTAD}
Victor Sanh, Lysandre Debut, Julien Chaumond, and Thomas Wolf. 2019.
\newblock Distilbert, a distilled version of bert: smaller, faster, cheaper and
  lighter.
\newblock \emph{ArXiv}, abs/1910.01108.

\bibitem[{Singhal(2001)}]{3320}
Amit Singhal. 2001.
\newblock Modern information retrieval: A brief overview.
\newblock \emph{IEEE Data Eng. Bull.}, 24:35--43.

\bibitem[{Tru{\c{s}}c{\u{a}}(2019)}]{trucscua2019efficiency}
Maria~Mihaela Tru{\c{s}}c{\u{a}}. 2019.
\newblock Efficiency of svm classifier with word2vec and doc2vec models.
\newblock In \emph{Proceedings of the International Conference on Applied
  Statistics}, volume~1, pages 496--503.

\bibitem[{Vaswani et~al.(2017)Vaswani, Shazeer, Parmar, Uszkoreit, Jones,
  Gomez, Kaiser, and Polosukhin}]{vaswani2017attention}
Ashish Vaswani, Noam Shazeer, Niki Parmar, Jakob Uszkoreit, Llion Jones,
  Aidan~N Gomez, {\L}ukasz Kaiser, and Illia Polosukhin. 2017.
\newblock Attention is all you need.
\newblock In \emph{Advances in neural information processing systems}, pages
  5998--6008.

\bibitem[{Wolf et~al.(2019)Wolf, Debut, Sanh, Chaumond, Delangue, Moi, Cistac,
  Rault, Louf, Funtowicz et~al.}]{wolf2019huggingface}
Thomas Wolf, Lysandre Debut, Victor Sanh, Julien Chaumond, Clement Delangue,
  Anthony Moi, Pierric Cistac, Tim Rault, R{\'e}mi Louf, Morgan Funtowicz,
  et~al. 2019.
\newblock Huggingface's transformers: State-of-the-art natural language
  processing.
\newblock \emph{arXiv preprint arXiv:1910.03771}.

\bibitem[{Worsham and Kalita(2018)}]{worsham-kalita-2018-genre}
Joseph Worsham and Jugal Kalita. 2018.
\newblock \href {https://www.aclweb.org/anthology/C18-1167} {Genre
  identification and the compositional effect of genre in literature}.
\newblock In \emph{Proceedings of the 27th International Conference on
  Computational Linguistics}, pages 1963--1973, Santa Fe, New Mexico, USA.
  Association for Computational Linguistics.

\bibitem[{Yang et~al.(2016)Yang, Lo, Xia, Bao, and Sun}]{7774514}
Xinli Yang, David Lo, Xin Xia, Lingfeng Bao, and Jianling Sun. 2016.
\newblock \href {https://doi.org/10.1109/ISSRE.2016.33} {Combining word
  embedding with information retrieval to recommend similar bug reports}.
\newblock In \emph{2016 IEEE 27th International Symposium on Software
  Reliability Engineering (ISSRE)}, pages 127--137.

\bibitem[{Yang et~al.(2019{\natexlab{a}})Yang, Cer, Ahmad, Guo, Law, Constant,
  Abrego, Yuan, Tar, Sung, Strope, and Kurzweil}]{yang2019multilingual}
Yinfei Yang, Daniel Cer, Amin Ahmad, Mandy Guo, Jax Law, Noah Constant,
  Gustavo~Hernandez Abrego, Steve Yuan, Chris Tar, Yun-Hsuan Sung, Brian
  Strope, and Ray Kurzweil. 2019{\natexlab{a}}.
\newblock \href {http://arxiv.org/abs/1907.04307} {Multilingual universal
  sentence encoder for semantic retrieval}.

\bibitem[{Yang et~al.(2019{\natexlab{b}})Yang, Dai, Yang, Carbonell,
  Salakhutdinov, and Le}]{yang2019xlnet}
Zhilin Yang, Zihang Dai, Yiming Yang, Jaime Carbonell, Russ~R Salakhutdinov,
  and Quoc~V Le. 2019{\natexlab{b}}.
\newblock Xlnet: Generalized autoregressive pretraining for language
  understanding.
\newblock \emph{Advances in neural information processing systems}, 32.

\end{thebibliography}
\bibliographystyle{acl_natbib}

\end{document}